\documentclass[journal,twoside,web]{IEEEtran}
\usepackage{cite}
\usepackage{amsmath,amssymb,amsfonts}
\usepackage{graphicx}
\usepackage{hyperref}
\usepackage{textcomp}
\usepackage{subcaption}  % For subfigures
\usepackage{bm}
\usepackage{bbm}
\usepackage[T1]{fontenc}
\usepackage{placeins}
\usepackage{algorithm2e} % added
\usepackage{comment}
\SetKwInput{KwInput}{Input}                % Set the Input
\SetKwInput{KwOutput}{Output}              % set the Output
\newcommand{\R}{{\mathbb R}}
\newcommand{\Mesh}{\mathcal{M}}
\newcommand{\Vertices}{\mathcal{V}}
\newcommand{\Points}{\mathcal{P}}

\newcommand{\Camera}{\mathcal{C}}
\newcommand{\Focus}{\mathcal{S}}
\newcommand{\FFocus}{{\tilde{\Focus}}}
\newcommand{\pphi}{{\tilde{\phi}}}

\newcommand*\diff{\mathop{}\!\mathrm{d}}
\DeclareMathOperator*{\argmax}{arg\,max}
\DeclareMathOperator*{\argmin}{arg\,min}

\def\BibTeX{{\rm B\kern-.05em{\sc i\kern-.025em b}\kern-.08em
    T\kern-.1667em\lower.7ex\hbox{E}\kern-.125emX}}
\begin{document}
\title{A Shape-Aware Total Body Photography System for In-focus Surface Coverage Optimization}
\author{Wei-Lun Huang, Joshua Liu, Davood Tashayyod, Jun Kang, Amir Gandjbakhche, Misha Kazhdan, Mehran Armand
\thanks{
% Manuscript submitted September 30, 2024.
The research was in part supported by the Intramural Research Program (IRP) of the NIH/NICHD, Phase I of NSF STTR grant 2127051, Phase II of NSF STTR grant 2335086, and Phase I NIH/NIBIB STTR grant R41EB032304.}
\thanks{Wei-Lun Huang is a Ph.D student in the Department of Computer Science,
        Johns Hopkins University, Baltimore, MD, USA and a pre-doctoral visiting fellow at the Eunice Kennedy Shriver National Institute of Child Health and Human Development, Bethesda, MD, USA (e-mail: whuang44@jhu.edu)}
\thanks{Joshua Liu is a research engineer at the Institute for Integrative and Innovative Research (I3R) at the University of Arkansas, USA (e-mail: joshua.liu@uark.edu)}%
\thanks{Davood Tashayyod is the CEO of Lumo Imaging. (e-mail: davood@lumoscan.com)}%
\thanks{Jun Kang is with the Faculty of the Department of Dermatology, Johns Hopkins School of Medicine, Baltimore, MD, USA (e-mail: jkang60@jh.edu)}%
\thanks{Amir Gandjbakhche is the Principal Investigator in Section on Translational Biophotonics at the Eunice Kennedy Shriver National Institute of Child Health and Human Development, Bethesda, MD, USA (e-mail: gandjbaa@mail.nih.gov)}%
\thanks{Misha Kazhdan is with the Faculty of the Department of Computer Science, Johns Hopkins University, Baltimore, MD, USA (e-mail: misha@cs.jhu.edu)}%
\thanks{Mehran Armand is with the Faculty of Institute for Integrative and Innovative Research and the Department of Mechanical Engineering, University of Arkansas, Fayetteville, AR, USA (e-mail: marmand@uark.edu)}%
}

\maketitle

\begin{abstract}
Total Body Photography (TBP) is becoming a useful screening tool for patients at high risk for skin cancer. While much progress has been made, existing TBP systems can be further improved for automatic detection and analysis of suspicious skin lesions, which is in part related to the resolution and sharpness of acquired images. This paper proposes a novel shape-aware TBP system automatically capturing full-body images while optimizing image quality in terms of resolution and sharpness over the body surface. The system uses depth and RGB cameras mounted on a 360-degree rotary beam, along with 3D body shape estimation and an in-focus surface optimization method to select the optimal focus distance for each camera pose. This allows for optimizing the focused coverage over the complex 3D geometry of the human body given the calibrated camera poses. We evaluate the effectiveness of the system in capturing high-fidelity body images. The proposed system achieves an average resolution of 0.068 mm/pixel and 0.0566 mm/pixel with approximately 85\% and 95\% of surface area in-focus, evaluated on simulation data of diverse body shapes and poses as well as a real scan of a mannequin respectively. Furthermore, the proposed shape-aware focus method outperforms existing focus protocols (e.g. auto-focus). We believe the high-fidelity imaging enabled by the proposed system will improve automated skin lesion analysis for skin cancer screening.
\end{abstract}

\begin{IEEEkeywords}
Total body photography, Skin cancer screening, Multi-view scanning, Extended depth of field
\end{IEEEkeywords}

\section{Introduction}
\label{sec:introduction}
% intro to TBP
\IEEEPARstart{T}{otal} Body Photography (TBP), a method that photographs the entire cutaneous surface of a person through a series of images of various body sectors, is becoming commercially available to aid in early detection of skin cancer in high-risk individuals \cite{halpern2003total}. TBP allows monitoring of temporal changes in skin lesions, efficient screening of numerous lesions, and recording anatomical locations for dermatoscopic images \cite{deinlein2020importance,hornung2021value,primiero2024narrative}. 

% the importance of improving the image quality in TBP 
In TBP imaging, image quality can be affected by artifacts from reflections, image compression, motion blur, lighting, and shadows, imposing challenges to identifying suspicious skin lesions \cite{dugonik2020image}. While existing TBP systems \cite{korotkov2014new,strzelecki2021skin,ahmedt2023monitoring} may tackle the mentioned issues, further increasing the image quality in terms of resolution and sharpness is an important step towards improving the sensitivity and specificity of automatic lesion classification and detection. For example, highly accurate and skilled clinical detection of melanoma relies heavily on established visual patterns and comparative recognition of suspicious lesions. Intuitive visual patterns include the well-known ABCDE criteria of melanoma (Asymmetry, Border irregularity, Color irregularity, Diameter>6mm, Evolving) \cite{nachbar1994abcd} and the 7-point checklist (identifying seven dermatoscopic criteria including pigment network, blue whitish veil, vascular structures, pigmentation, streaks, dots and globules, and regression structures) \cite{argenziano1998epiluminescence}. Comparative recognition of suspicious lesions relies on “ugly duckling” recognition (a nevus that is obviously different from the others in a given individual) \cite{gaudy2017ugly}. Since TBP provides an efficient way to screen a large number of lesions with its wide-view pictures, it can also be used for detecting lesion saliency for assessing suspicious skin lesions \cite{mohseni2021can,yu2021end,soenksen2021using}. Mohseni \textit{et al.} \cite{mohseni2021can} suggest a lesion should occupy an area of at least 20$\times$20 pixels to be reliably detected by algorithms and for automated analysis. As a result, a 1.5 mm diameter lesion requires a resolution of around 0.075 mm/pixel, whereas smaller lesions cannot be reliably detected or analyzed with a 10 MP smartphone camera capturing an image at about 20 cm from the subject \cite{soenksen2021using}. Therefore, improving image quality (resolution and sharpness) will significantly benefit suspicious lesion detection.

% challenges
Camera focus remains a challenge for improving resolution and sharpness in a TBP system for several reasons. First, to reach high pixel resolution per unit surface area (defined as equivalent system resolution hereafter), a TBP system needs to use either large focal length lenses or close subject-to-camera distances. In both cases, the sharpness of images is limited because the camera gives a shallow depth of field -- the distance between the closest and the furthest objects that appear at acceptable sharpness in an image.
Second, Auto-focus (AF), though well-studied for single-image capture and supported in most modern cameras, is greedy in the context of multi-view capture like TBP. AF does not consider if a given point has already been seen by some other camera \cite{abuolaim2018revisiting,abuolaim2020online} and may fail when imaging textureless objects. Third, camera focus needs to adapt to human shapes and poses, especially since no standardized pose exists for TBP yet\cite{halpern2003standardized}. Using a set of pre-defined focus distances in a TBP system is susceptible to out-of-focus images when scanning individuals with various body shapes and in different poses. Finally, extending the depth of field by focus stacking or focal sweeping \cite{kuthirummal2010flexible,hausler1972method} is infeasible for a TBP system with moving cameras and when scanning time is restricted.

% proposed method
In this paper, we first formulate TBP scanning as an image quality (resolution and sharpness) optimization problem over the body surface. In the optimization problem, assuming the camera poses and the 3D mesh of the subject are given, we solve for a focus distance for each camera pose using the Expectation-Minimization (EM) algorithm, dubbed "Shape-aware Focus". The goal is to optimize the quality of the imaged surface, cumulatively seen from multiple 2D views of a subject. The paper presents a novel shape-aware TBP system and an imaging pipeline. We evaluate the performance of individual components in the proposed pipeline and demonstrate the effectiveness of the system in capturing high-quality images of the entire body. We believe that the proposed TBP system can efficiently and effectively image the patient's skin and corresponding lesions, thus improving downstream tasks in skin cancer screening such as suspicious lesion detection and longitudinal tracking of lesions. 

% contributions
The three main contributions of this paper include:
\begin{enumerate}
    \item We formulate the TBP scanning problem as a multi-view capture when restricted to determining the camera's depth of field and propose the use of the EM method that optimizes focus distances.
    \item We perform systematic evaluations of the proposed TBP image capture protocol, including system calibration, 3D shape estimation, and shape-aware focus. 
    \item As a proof-of-concept, we prototype the shape-aware TBP system, validate the system on simulation data to show the superiority of the shape-aware focusing method and its generalizability to diverse body shapes and poses. We also validate the system using a real scan mannequin to demonstrate the proposed method outperforms existing image capture protocols (i.e. auto-focus).
\end{enumerate}

\section{RELATED WORKS}
\subsection{Total Body Photography System for Skin Cancer}
%commercial system
Many specialized systems have been developed for TBP of skin cancer. Some commercialized TBP systems are available in the US market such as Canfield Vectra 360 \cite{rayner2018clinical}, ATBM master Fotofinder Systems GmbH \cite{winkler2024performance}, and MelanoScan \cite{drugge2009melanoma}.
%TBP system 2D
TBP systems can be broadly categorized into image-based and 3D-mesh-based systems. Among image-based systems, Dengel \textit{et al.} \cite{dengel2015total} prototyped a rapid TBP imaging system with 16 cameras to acquire multiple photos of patient skin in two poses for complete coverage. Korotkov \textit{et al.} \cite{korotkov2014new} designed a more complex system comprising 21 high-resolution cameras and a turntable, demonstrating capabilities for automatic mapping of pigmented skin lesions \cite{korotkov2018improved} and estimating lesion changes using sets of overlapping images. Strzelecki \textit{et al.} \cite{strzelecki2021skin} developed a TBP system with a single digital camera that rotates and moves vertically around the subject. The camera captures 32 images for lesion detection and lesion matching \cite{strkakowska2022skin}. Addressing the challenge of accurate lesion measurement due to unknown camera distances and orientation angles, Szczypinski \textit{et al.} \cite{szczypinski2021orthorectification} created a TBP system using four 20-MP-resolution cameras and two depth sensors to orthorectify RGB images to enable the correction of new geometries.

%TBP system 3D
Although 2D image-based systems generally preserve sharper texture, 3D-mesh-based systems offer robust monitoring of temporal changes in lesions, accommodating variations in camera configurations and patient body poses.
Bogo \textit{et al.}\cite{bogo2014automated} developed a multi-camera 3D stereo system for detecting new lesions or the growth of existing ones. Zhao \textit{et al.} \cite{zhao2022skin3d} and Huang \textit{et al.} \cite{huang2023skin} proposed methods for longitudinal tracking of skin lesions using 3D textured meshes captured by the Artec Shapify Booth 3D full-body scanner \cite{saint20183dbodytex}. Recently, Ahmedt-Aristizabal \textit{et al.} \cite{ahmedt2023monitoring} developed a 3D whole-body imaging system that consists of sixty 24-MP-resolution cameras for skin screening.

%Limitations: image resolution and limited depth of field issues
Despite these advancements, existing TBP systems face limitations in image sharpness and resolution. For instance, the system described by Korotkov \textit{et al.} \cite{korotkov2014new} uses a fixed focus distance of approximately 400-450 mm for each camera, resulting in blurry images of body areas that deviate significantly from this distance. For example, due to constrained depth of field, when a camera is capturing from the lateral side of a patient, the thigh closer to the camera will be in focus while the other thigh will not be. Furthermore, while Ahmedt-Aristizabal \textit{et al.} \cite{ahmedt2023monitoring} utilizes autofocus in their system, the reported resolution is approximately 0.08 mm/pixel, insufficient to stably detect lesions smaller than 2 mm in diameter. We note that most of the other studies do not report the equivalent system resolution. To address these limitations, our system adopts short subject-to-camera distances for higher system resolution and autonomously selects an optimal focus distance for each camera to optimize the focused surface area.
\subsection{Extended depth of field}
Extending depth of field or multi-focus image fusion, achieved by capturing multiple images at different focal planes followed by image fusion is a technique for acquiring an all-in-focus image of a target object \cite{liu2020multi,kuthirummal2010flexible}.
Str{\"o}bel \textit{et al.} \cite{strobel2018automated} proposed an automated device to combine extended depth of field images from multiple views to reconstruct 3D models of pinned insects and other small objects. For extended depth of field in multi-view images, Chowdhury \textit{et al.} \cite{chowdhury2021fixed} proposed a fixed-lens multi-focus image capture technique and calibrated image registration to mitigate artifacts in the fused images due to violation of perspective image formation. However, in multi-focus image fusion applications, the required number of source images is usually undetermined without depth information. Recently, Skuka \textit{et al.} \cite{skuka2022extending} proposed to extend the depth of field of the imaging systems based on the depth map of the scene. However, they only focus on a single target scene with the same camera pose across all images. Though the approach extends to multiple camera poses, the computational complexity is exponential in the number of poses for cameras with overlapping fields of view. In total body photography, long scanning time is not ideal for human subjects. In this paper, we formulate the scanning problem as constrained to the camera's depth of field and solve it efficiently to extend the depth of field of the proposed system. 

\section{METHOD}
\subsection{Problem Formulation and Cost Function}
We formulate the TBP scanning problem as follows: Given a mesh $\Mesh$ and camera poses $\Camera\subset\mathrm{SE}(3)$, we would like to solve for an assignment of focus distances $\Focus \in \R^{|\Camera|}$ to cameras that reduces the area of poorly imaged surface. We formulate this by (1)~assigning a per-camera cost to each point on the surface, (2)~defining the cost per-point as the minimum cost over all cameras, and (3)~seeking the focus distances minimizing the integrated cost over all surface points.

\subsubsection{Focus Distance Cost}
\begin{figure}
  \centering
  \begin{subfigure}{0.35\textwidth}
    \includegraphics[width=\textwidth]{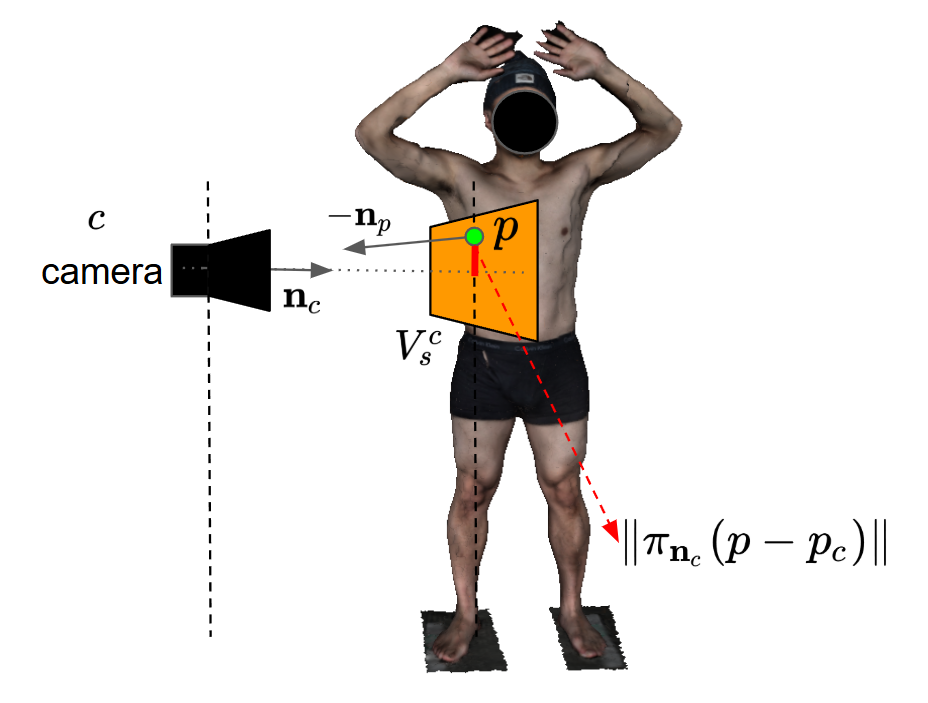}
    \caption{Cost-determining factors}
    \label{fig:criteria}
  \end{subfigure}
  \hfill
  \begin{subfigure}{0.35\textwidth}
    \includegraphics[width=\textwidth]{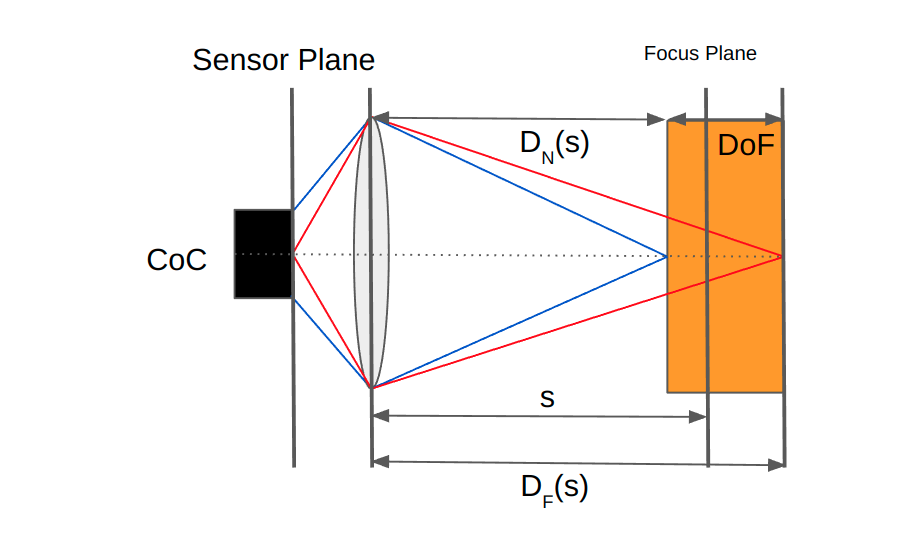}
    \caption{The principle of the depth of field}
    \label{fig:dof}
  \end{subfigure}
  \caption{Visualization for the cost-determining factors and the principle of the depth of field. In (a), the orange frustum ($V^c_s\subset\R^3$) is clipped at the near and far depth of field (DoF) limits. The red line represents the deviation from the optical axis. In (b), $CoC$ is the circle of confusion, $s$ is the focus distance, and $D_N(s)$ and $D_F(s)$ are the near and far depth of field limits.}
  \label{fig:criteria_and_dof}
\end{figure}

Given a camera $c\in\Camera$ and focus distance $s\in\R$, we define a cost function $\kappa_s^c:\Mesh\rightarrow[0,1]$. The function is set to one (the maximum cost) for all surface points $p\in\Mesh$ that are invisible to camera $c$. Otherwise, the cost for a given point is determined by the projected area on the image plane, its deviation from the optical axis due to field curvature~\cite{matsunaga2015field}, and the proximity to the focal plane:
\begin{equation}
\begin{aligned}
\kappa^c_s(p)
= {}& w_1 \cdot \min\underbrace{\left(\tfrac{\varepsilon_1\langle p-p_c,\vec{n}_c\rangle^2}{\langle\vec{n}_c,\vec{n}_p\rangle}, 1\right)}_{\hbox{\tiny projected area}} \\
{}+{}& w_2 \cdot \min\underbrace{\left(\tfrac{\|\pi_{\vec{n}_c}(p-p_c)\|}{\varepsilon_2}, 1 \right)}_{\hbox{\tiny optical axis deviation}} \\
{}+{}& w_3 \cdot \underbrace{\left(1 - \mathbbm{1}(p \in V^c_s)\right)}_{\hbox{\tiny proximity to focal plane}}
\label{eqn:cost function}
\end{aligned}
\end{equation}
where
\begin{itemize}  
\item $w_i\in\R^{\geq0}$ is the weight for the $i$-th cost term,
\item $\varepsilon_i\in\R^{\geq0}$ are thresholding values,
\item $p_c \in \R^3$ is the position of camera $c$,
\item $\vec{n}_p\in S^2$ is the surface normal at $p$,
\item $\vec{n}_c\in S^2$ is the viewing direction of camera $c$,
\item $\pi_{\vec{n}}:\R^3\rightarrow\R^3$ is the projection onto the plane perpendicular to $\vec{n}$,
\item $\mathbbm{1}(\cdot)$ is an indicator function, equal to one if the condition is true and zero otherwise, and
\item $V^c_s\subset\R^3$ is the view frustum of camera $c$ with the near and far clipping planes set to the near and far depth of field limits (Fig.~\ref{fig:dof}) for a focus distance $s$.
\end{itemize}
In the projected area term, the projection area of a 3D patch is a function of the depth (the distance along the optical axis of the camera) and incidence (the alignment between the viewing direction and the surface normal). Note that we define the surface normals as inward-pointing. $\varepsilon_1$ is a threshold for the projected area so that the projected area term approaches zero for an infinite projected area and goes to 1 for a zero projected area. In the optical axis deviation term, the deviation is defined by the distance between the projection of the point $p$ and the image center on the image plane. $\varepsilon_2$ is a threshold so that the optical axis deviation term equals zero for a point projected onto the image center and approaches 1 for infinite deviation. We scale and clamp individual terms to the range $[0, 1]$ and use equal weights, $w_i=1/3$. A visualization of the cost-determining factors is shown in Fig.~\ref{fig:criteria}.

\subsubsection{Total Focus Distance Cost}
We define the total focus distance cost, $\mathcal{K}:\R^{|\Camera|}\rightarrow\R^{\geq0}$, by integrating the minimal pointwise cost over all cameras, over all surface points:
\begin{equation}
\label{eqn:total focal cost}
    \mathcal{K}(\Focus) = \int_{\Mesh} \kappa_\Focus(p) \diff p\quad\hbox{w/}\quad
    \kappa_\Focus(p)=\min_{c\in\Camera}\kappa_{\Focus_c}^c(p)
\end{equation}
The solution is the set of focus distances minimizing the cost:
\begin{equation}
\label{eqn:optimization}
\Focus = \argmin_{\FFocus\in\R^{|\Camera|}}\left(\mathcal{K}(\FFocus) =\int_\Mesh\kappa_\FFocus(p)\diff p\right).
\end{equation}

\subsection{Shape-aware Focus}
\label{ss:shape_aware_focus}
We use Expectation-Minimization (EM) to solve the optimization in \eqref{eqn:optimization}, called "Shape-aware Focus". The proposed method determines optimal focus distances that adapt to the input shape, addressing the challenge that the strategy for selecting focus distances should accommodate patients of diverse body shapes and poses. 
Concretely, the optimization in \eqref{eqn:optimization} can be expressed as a simultaneous optimization over assignments of surface positions to cameras, $\phi:\Mesh\rightarrow\Camera$, and focus distances, producing the EM problem:
\begin{equation}
(\phi,\Focus) = \argmin_{(\pphi,\FFocus)}\left(\mathcal{K}(\pphi,\FFocus)=\int_\Mesh\kappa_{\FFocus_{\pphi(p)}}^{\pphi(p)}(p)\diff p\right).
\label{eqn:em_optimization}
\end{equation}
The equivalence follows from the fact that any set of focus distances implicitly defines an assignment of points to cameras, with a point assigned to the camera minimizing the point's cost. In practice, we approximate using Monte-Carlo integration. Letting $\Points\subset\Mesh$ be a discrete point-set, we set:
\begin{equation}
(\phi,\Focus)=\argmin_{(\pphi,\FFocus)}\sum_{p\in\Points}\kappa_{\FFocus_{\pphi{p}}}^{\pphi(p)}(p).    
\end{equation}
We use the EM approach to compute the assignment $\phi$ and focus distances $\Focus$ (in Algorithm~\ref{alg:em}) by first initializing the focus distances and then alternately fixing the focus distances and solving for the assignments (Step 1), and fixing the assignments and solving for the focus distances (Step 2).
\begin{algorithm}
 \caption{Expectation-Minimization algorithm}
    \label{alg:em}
    \SetKwFor{Loop}{Loop}{}{EndLoop}
    \SetKwInOut{KwIn}{Input}
    \SetKwInOut{KwOut}{Output}
    \KwIn{$\Points$, $\Camera$ }
    \KwOut{$\phi$, $\Focus$}
    
    Initialize $\Focus$\\
    $\mathcal{K}_{\mathrm{old}} = \infty$\\
    \Loop{}{
    1. Solve for the assignment $\phi$ minimizing:
        \begin{equation}
            \phi \leftarrow \argmin_{\pphi}\left(\mathcal{K}_\Focus(\pphi) = \sum_{p \in \Points} \kappa_{\Focus_{\pphi(p)}}^{\pphi(p)}(p)\right)
        \end{equation}
    2. Solve for focus distances $\Focus$ minimizing:
        \begin{equation}
            \Focus \leftarrow \argmin_{\FFocus}\left(\mathcal{K}_\phi(\FFocus) = \sum_{p \in \Points} \kappa_{\FFocus_{\phi(p)}}^{\phi(p)}(p)\right)
        \end{equation}

    \If{$\mathcal{K}(\phi,\Focus)>\mathcal{K}_{\mathrm{old}}\cdot(1-\varepsilon)$}{break}
    $\mathcal{K}_{\mathrm{old}} = \mathcal{K}(\phi,\Focus)$
    }
\end{algorithm}

\subsubsection{Assignment step}
In the assignment step, we solve for the function $\phi: \Points \rightarrow \Camera$ given estimated focus distances $\Focus$. That is, we simply assign a point to the camera minimizing the cost for that point:
\begin{equation}
\label{eqn:assignment}
\phi(p) = \argmin_{c\in\Camera}\kappa_{\Focus_c}^c(p).
\end{equation}

\subsubsection{Minimization step}
\label{ss:minimization_step}
In the minimization step, we solve for focus distances $\Focus \in \R^{|\Camera|}$ given the estimated assignment $\phi: \Points\rightarrow \Camera$. As the assignments are fixed, this is done independently for each camera, with the focus distance being the value minimizing the contribution from the assigned points.
\begin{equation}
\label{eqn:minimization}
\Focus_c = \argmin_{s\in\R} \sum_{p \in\phi^{-1}(c)} \kappa_s^c(p)
\end{equation}
For a given camera $c$, using the fact that the cost from the projected area and the optical axis deviation are independent of focus distance, this reduces \eqref{eqn:minimization} to:
\begin{equation}
\label{eqn:minimization reduced}
\Focus_c = \argmax_{s\in\R}\sum_{p\in\phi^{-1}(c)}\mathbbm{1}(p \in V^c_s)
\end{equation}
That is, the optimal focus distance $\Focus_c$ is the distance at which the largest subset of points assigned to camera $c$ are in its view-frustum. Since the summation in \eqref{eqn:minimization reduced} is piecewise constant in $s$, we can find an optimal focus distance by partitioning the range of focus distances into intervals over which the summation is constant. Then, finding the optimal focus distance reduces to finding the interval over which the number of in-frustum points is maximized.

\subsection{System Design and Imaging Pipeline}
\begin{figure*}[thpb]
    \centering
  \includegraphics[width=0.85\textwidth]{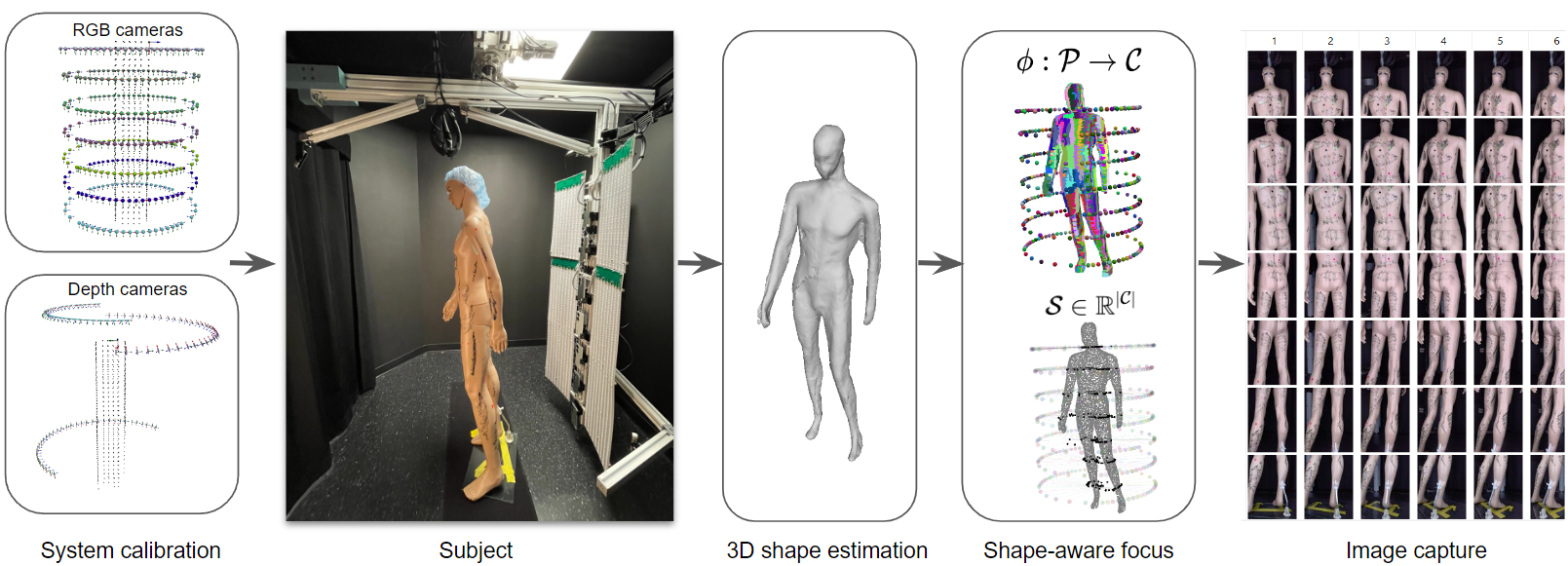}
  \caption{Proposed pipeline. The system calibration is performed to acquire the poses of high-resolution RGB cameras and depth cameras. The subject stands at the designated position. The system captures multiple depth images by rotating 180 degree to estimate the 3D shape of the subject through TSDF-based method. The system selects a focus distance for each RGB camera pose using the proposed shape-aware focus method to optimize the quality of surface coverage. Finally, the system rotates around the subject and takes pictures.}
  \label{fig:pipeline}
\end{figure*}

\begin{figure}[thpb]
      \centering
      \includegraphics[height=6cm]{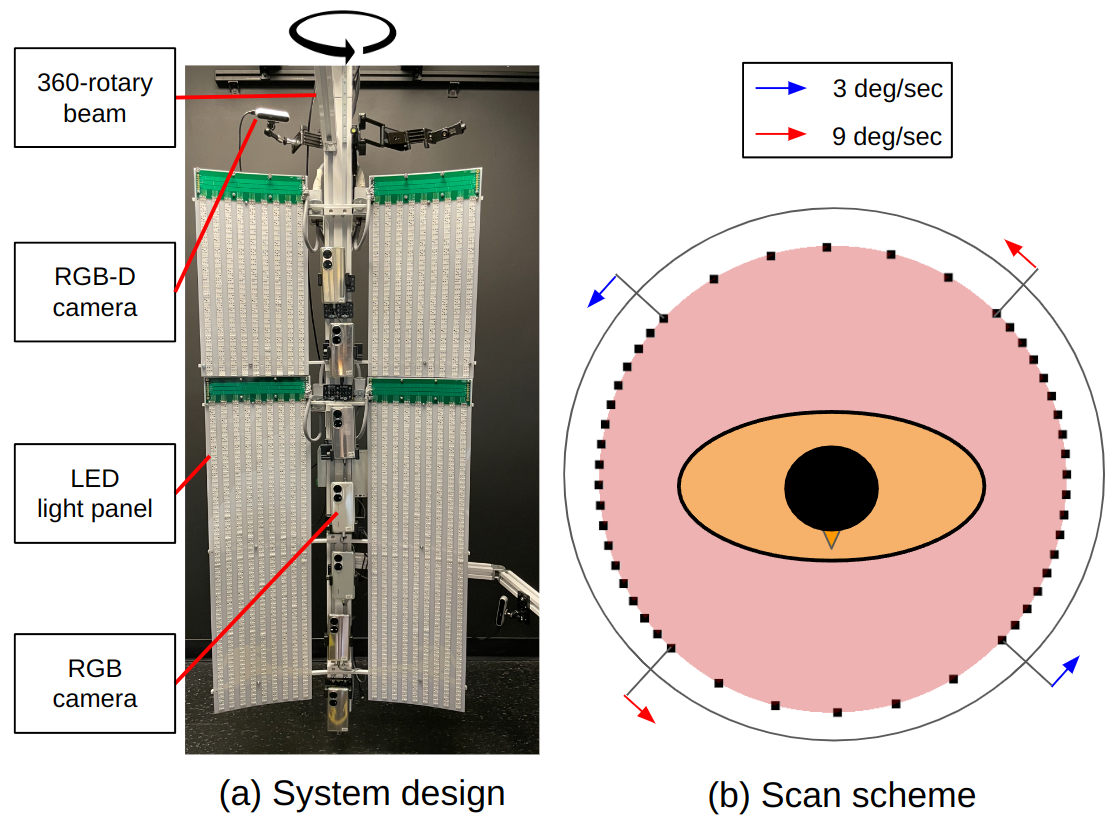}
      \caption{Illustration of the (a) proposed shape-aware imaging system and (b) the schematic top-view of the acquisition in a scan.}
      \label{shape aware system}
\end{figure}

We propose a new imaging pipeline (Fig.~\ref{fig:pipeline}) and design the shape-aware TBP system with depth and RGB cameras on a 360-degree rotary beam (Fig.~\ref{shape aware system}(a)). Since the problem formulation requires camera poses and the 3D mesh to be given, we perform system calibration to estimate a set of poses for RGB cameras, followed by the 3D shape estimation of the subject from depth cameras. Subsequently, based on shape-aware focus, a focus distance is selected for each RGB camera pose to capture images that optimize in-focus surface coverage. Finally, the rotary beam rotates around the subject and the cameras capture images at designated poses.

The shape-aware TBP system consists of three depth cameras and seven 48-MP resolution RGB cameras on a 360-degree rotary beam, as shown in Fig.~\ref{shape aware system}a. The schematic top-view of the acquisition in a scan for one RGB camera is visualized in Fig.~\ref{shape aware system}b. The LED light panel is used to set the ISO value of the RGB cameras to the lowest value, thereby reducing noise. Two depth cameras are installed at the side of RGB cameras, and the third depth camera is installed at the opposite side so that we can collect depth images faster to estimate the object's shape. To produce a 3D reconstruction with high-resolution textures, the RGB cameras are aligned vertically with 30\%-50\% overlap in the field of view for adjacent cameras. Also, each RGB camera capture images at 48 angular positions along a circular trajectory. We notice that the variation in depth is larger at the lateral regions in the designated pose, therefore, we position the cameras sparsely at the anterior and posterior regions and more densely in the lateral regions. Such an angular distribution of cameras coincides with \cite{korotkov2014new} and is also beneficial for 3D reconstruction. We set two rotational speeds in a scan, 9$^\circ$/sec when the system is at the front/back regions and 3$^\circ$/sec for the lateral regions. The focus distance for a camera is determined for each angular position before the scan, allowing us to adjust the focus distance as the system rotates, without having to pause the acquisition.

\subsection{System Calibration}
Since the scanning problem assumes camera poses are given, we need to estimate a set of poses for RGB cameras denoted as $\Camera\subset\mathrm{SE}(3)$. To achieve this, we first estimate intrinsic camera parameters (focal length and principal point) and distortion parameters via the well-established method from Chang \cite{zhang2000flexible}, denoted as $\theta$. Additionally, we design a calibration cuboid with dimensions 280$\times$280$\times$1800 (mm) motivated by \cite{an2018ChArUco}.
The cuboid consists of 4 distinct ChArUco \cite{garrido2014automatic} boards. The calibration cuboid is suitable for calibrating 360-degree inward-facing cameras in two folds: 1) The calibration process is well-estimated with sufficient calibration data in the subject-occupied regions, the same as the workspace of the calibration objects. 2) The ChArUco-board allows for partial visibility of the pattern in captured images and provides a unique and accurate 2D-3D correspondence.

We first detect the checkerboard corners of the ChArUco boards as keypoints in the images. We create a list of image pairs. In each image pair, we then specify the keypoint correspondences when the two images share common keypoints in the views. We perform calibration using incremental Structure from Motion (SfM) \cite{moulon2013adaptive,schonberger2018robust}. To adjust the scale of the SfM reconstruction, we estimate a similarity transformation to bring the calibration results into the real scale. To achieve this, we rely on one of the ChArUco board patterns, with a known structure and the physical size of the grids. Let $\Points_{c}$ be the target points defined in the ChArUco board coordinate and $\Points_{S\!f\!M}$ be its correspondence defined in an arbitrary coordinate system from the SfM. We would like to find the transformation matrix $H$ composed of a uniform scale $s$, a rotation matrix $R \in SO(3)$, and a translation vector $\Vec{t} \in R^3$:
\begin{equation}
    H(s, R, \Vec{t}) = \argmin_{s, R, \Vec{t}} \sum_i^N || \Points_{c}^i - H(s, R, \Vec{t})\Points_{S\!f\!M}^i ||^2,
\end{equation}
where $N$ is the number of keypoints on the ChArUco board.

\subsection{3D Shape Estimation}
Since the problem formulation assumes the 3D mesh to be given, we need to estimate the 3D shape of the subject before a scan. We note that for non-static objects, e.g. human subjects, the run-time for the 3D shape estimation is a limiting factor. There is a higher chance of changes in body pose and dynamic wobbling with longer running-times, leading to a divergence of the estimated shape from the real shape over the course of the scan. We use two alternative methods for the 3D shape estimation of the subjects.
\subsubsection{Template-based mesh estimation}
We capture single-shot depth image from the 3 depth cameras, followed by processing with background removal, a noise filter, cropping within a pre-designed volume, computing the largest connected components, and stitching. However, the point cloud is incomplete and prevents us from determining appropriate focus distances. Several approaches have been proposed to register an SMPL \cite{loper2023smpl} model to a partial point cloud \cite{jang2023dynamic,marin24nicp}. We follow the approach from Marin \textit{et al.} \cite{marin24nicp}. A learned neural field dedicated to different parts of the shape is used to predict vertex displacement of the SMPL model. The parameters of the neural field are then refined using Iterative Closest Point \cite{besl1992method} through backpropagation. Then, the updated neural field is utilized to register the SMPL model to the input point cloud. Finally, the registration output by the network is further refined by an optimization using Chamfer distance. We denote by SMPL-NICP the method from \cite{marin24nicp}.
\subsubsection{Truncated Signed Distance Function reconstruction}
Alternatively, we acquire multiple depth images with each of the 3 depth cameras. Then we compute the truncated signed distance function (TSDF) \cite{curless1996volumetric} from the depth images and their poses, and optimize the signed distance function. However, the collected depth images are usually noisy. To reject outliers, a truncation function is applied to the output value of the signed distance function. Finally, for a sequence of inputs, per-point signed distances can be estimated using the weighted average from all the frames, where the weight depends on view angles and distances. We then apply the Ball Pivoting algorithm \cite{bernardini1999ball} to reconstruct a mesh given the TSDF-based point cloud.

\section{Implementation}
\label{s:implementation}
We use a Huawei P50 Pro and Realsense D415 as the high-resolution RGB cameras and the RGB-D cameras. The Huawei P50 Pro has a focal length of 6 mm, an aperture of f/1.8, and a sensor size of 1/1.55 inches, equivalent to a hyperfocal distance of 2,860 mm. We note that in reality, the image sharpness degrades when deviating from the focal plane. Therefore, in Sec.\ref{s:results}, we add a factor of 2 (i.e. 5720 mm) in the depth of field calculation for tighter control of the focus distance. The capturing configuration of the Huawei camera is set to an ISO speed of 50, shutter speed of 1/500 second, and white-balance of 3500 K. The image resolutions used for the system calibration are 8192 $\times$ 6144 (Huawei) and 1920 $\times$ 1080 (Realsense). The depth image resolution for TSDF is 1280 $\times$ 720. We use the manufactured transformation from the depth camera to the RGB camera in the Realsense D415 to close the transformation loop from the world frame (in calibration) to the depth camera's frame. For a complete TBP scan, we collect 336 high-resolution images in total.

\section{Results}
\label{s:results}
\subsection{System Calibration}
\subsubsection{Calibration accuracy}
We evaluate the single system calibration by measuring the standard reprojection error. We define a residual as the pixel-wise deviation between a 2D observation (as detected) of a keypoint and its projection (as estimated) in an image. For robust estimation of camera pose and 3D point triangulation, a residual with an error larger than 4 pixels is discarded in the Bundle Adjustment. The average reprojection error is 0.58 with around 30K residuals.
\subsubsection{System precision}
\label{ss:system precision}
For repeatability of camera poses, we register the estimated camera poses for a pair of system calibration results. For a pair of registered camera poses, $(t_1, R_1)$ and $(t_2, R_2)$, we compute the Euclidean distance $||t_1 - t_2||$ and the angle between rotation axes $\hbox{acos}\Big(\frac{\hbox{\footnotesize trace}(R_1^T, R_2) - 1}{2}\Big)$ as the translational and rotational error for the repeatability measure, where $t_i$ and $R_i$ are the translation vector and the rotation matrix of a camera pose. We perform system calibrations five times. The average translational and rotational error is 2.16 mm and 0.21 degrees.

\subsection{3D Shape Estimation}
\begin{figure}[thpb]
      \centering
      \includegraphics[height=4.5cm]{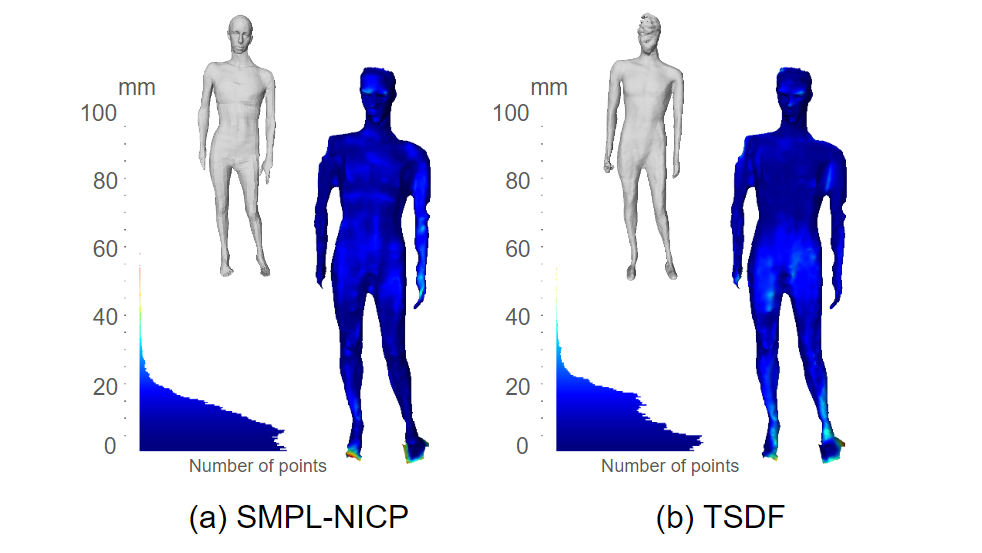}
      \caption{Visualization of the error for 3D shape estimation for (a) SMPL-NICP and (b) TSDF method. The white-gray mesh in (a) and (b) is the estimated shape. The GT mesh is colorized based on the error so that vertices with a smaller error are colored in dark blue and vertices with a larger error are colored in light colors. The histogram visualizes the distribution of the error for each method.}
      \label{f:shape evaluation}
\end{figure}

We use the 3D mesh from photogrammetry reconstruction \cite{moulon2017openmvg,cernea2020openmvs} as the ground truth. Since the reconstruction is correct up to a scale factor, we register the estimated camera poses from SfM to the calibrated camera pose to fix the scale. We compute the Chamfer distance from the ground-truth mesh to the estimated mesh as the quality measure:
\begin{equation}
    CH(\Vertices_0, \Vertices_1) =\frac{1}{|\Vertices_0|}\sum_{v_0\in\Vertices_0} \min_{v_1\in \Vertices_1} ||v_0-v_1||_2 ,
\end{equation}
where $\Vertices_0$ are the vertices of the ground-truth mesh, and $\Vertices_1$ are the vertices of the estimated mesh.
Fig.~\ref{f:shape evaluation} visualizes the distances (i.e. errors) for different 3D shape estimation methods. The Chamfer distance is 8.31 mm and 9.31 mm for the SMPL-NICP and the TSDF methods, respectively.
\subsection{Quality of Surface Coverage and Resulting Image}
\label{ss:eval shape aware focus}
\subsubsection{Metrics}
We compare the proposed method against two baseline approaches: focusing at the closest depth ("Closest Focus") and focusing at the average depth ("Average Focus"). The first metric is the total cost ($\mathcal{K}(\Focus)$) in \eqref{eqn:total focal cost}, as the accumulated cost per point (0 to 1) indicating if a point is well-captured. The second metric is the percentage of in-focus surface area, measured by calculating the proportion of sampling points within any camera's view frustum, denoted by $\Points(p \in V^c_s))$. Since the radius of the cylindrical camera network is approximately 450 mm, we established thresholds for the projection area and optical axis deviation at 2.47$\times10^{-6}$ mm$^2$ (equivalent to a distance of 450 mm with an incidence angle of 60 degrees) and 450 mm, respectively. We sample 10K points on the mesh for our analysis in both the simulation and the real scan.
Additionally, we report the visibility percentage of the surface to the TBP system and the system resolution (mm/pixel) across the surface. We note that both visibility percentage and system resolution are determined by the camera poses and 3D shape, and are independent of the focus distances.
\begin{figure}[thpb]
      \centering
      \includegraphics[height=4.5cm]{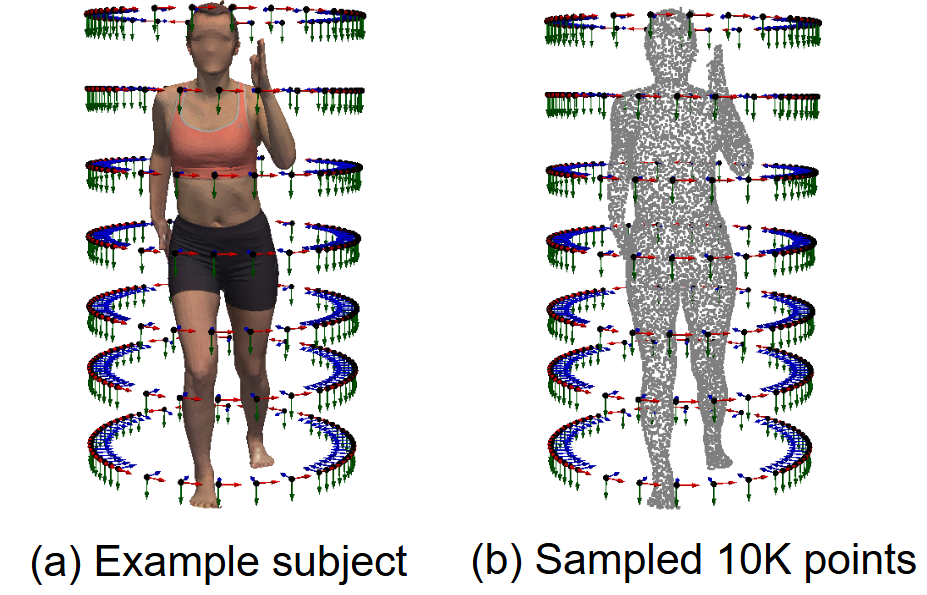}
      \caption{Visualization of (a) an example data from the 3DBodyTex dataset and (b) a set of uniformly sampled 10K points. The camera positions are visualized in black spheres, and the orientations are visualized in the red/green/blue-arrow for the x/y/z-axis.}
      \label{f:sample data}
\end{figure}
\subsubsection{Simulation Data}
We first evaluate the methods on the 3DBodyTex dataset \cite{saint20183dbodytex,saint2019bodyfitr}. The dataset consists of 400 textured 3D meshes of human subjects in various poses and shapes obtained from real scans. The average surface area for the entire dataset is 1,840,527 mm$^{2}$ with a standard deviation of 208,567 mm$^{2}$. Fig.~\ref{f:sample data} shows an example subject from the 3DBodyTex dataset and the sampled points of the subject. 
Table \ref{t:evaluation} shows that our approach improves the total cost by 17\% and 8.8\% as compared to "Closest Focus" and "Average Focus". Our approach also increases the in-focus area by 53.3\% and 16.4\% respectively as compared to "Closest Focus" and "Average Focus".
On average, 93.2\% ($\sigma=2.2\%$) of the surface is visible to the system.
The average system resolution across the surface is 0.068 mm/pixel (with $\sigma = $  0.007 mm/pixel across 400 meshes). On average, 73\% ($\sigma=5.2\%$) of the surface has a system resolution better than the 0.075 mm/pixel criterion across 400 meshes.
\begin{table}
    \centering
    \small
\begin{tabular}{|c|r|r|r|r|} 
\hline
Method & \multicolumn{1}{c|}{\textit{closest}} & \multicolumn{1}{c|}{\textit{average}} & 
\multicolumn{1}{c|}{\textit{shape-aware}}\\
\hline
\hline
$\mathcal{K}(\Focus)_{\text{sim}}$ $\downarrow$ & $4710\ (268)$ & $3871\ (325)$ & $2986\ (301)$ \\
\hline
$\Points(p \in V^c_s)_{\text{sim}}$ $\uparrow$ & $31.6\ (5.1)$ & $68.5\ (5.6)$ & $84.9\ (4.1)$ \\ 
\hline
\hline
$\mathcal{K}(\Focus)_{\text{real}}$ $\downarrow$ & $4541$ & $3507$ & $2818$ \\
% \hline
% $\mathcal{R}$ & $26.5$ & $20.3$ & $15.0$ \\
\hline
$\Points(p \in V^c_s)_{\text{real}}$ $\uparrow$ & $38.4$ & $80$ & $95.1$ \\ 
\hline
\end{tabular}
    \caption{Quantitative evaluation for the proposed methods. We note that the maximum possible total cost $\mathcal{K}(\Focus)$ is 10,000, occurring when all points have a cost value of 1. $\Points(p \in V^c_s)$ is the percentage of in-focus surface area. The subscript ("sim" and "real") denotes if the experiment is conducted on simulation or real scan data. For the simulation data, we also calculate the standard deviation (shown in the parenthesis) for the two metrics across 400 subjects.}
    \label{t:evaluation}
\end{table}
\subsubsection{Real TBP scan}
\begin{figure}
  \centering
  \begin{subfigure}{0.35\textwidth}
    \includegraphics[width=\linewidth]{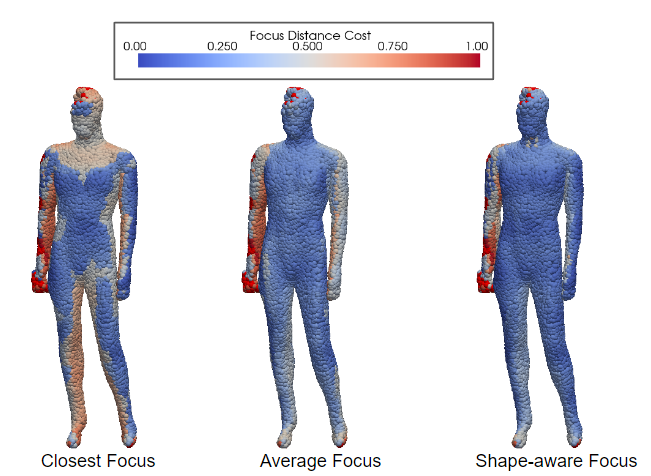}
    \caption{Pointwise cost over the surface}
    \label{fig:method_evaluation_1}
  \end{subfigure}
  \hfill
  \begin{subfigure}{0.35\textwidth}
    \includegraphics[width=\linewidth]{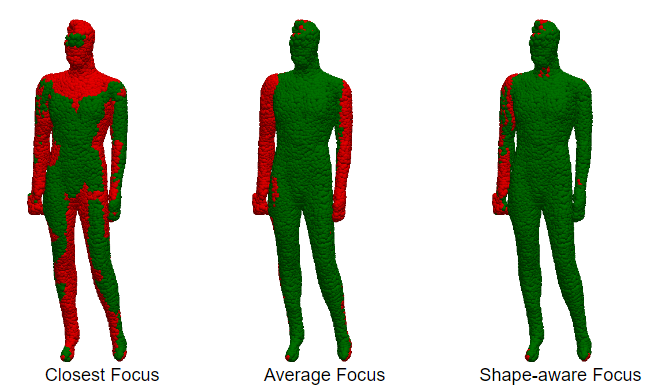}
    \caption{In-focus surface area}
    \label{fig:method_evaluation_2}
  \end{subfigure}
    \begin{subfigure}{0.35\textwidth}
    \includegraphics[width=\linewidth]{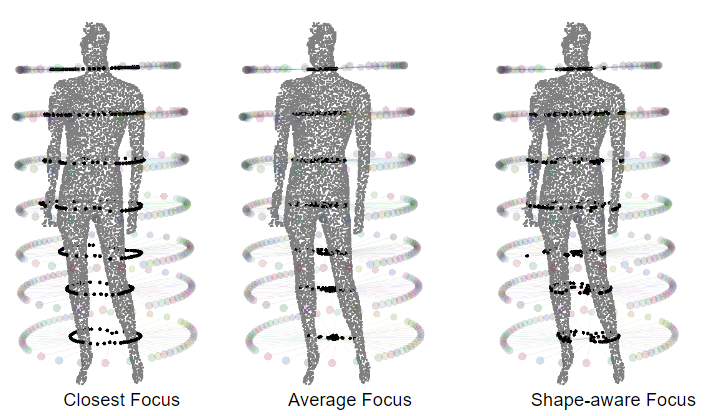}
    \caption{Selected focus distances}
    \label{fig:method_evaluation_3}
  \end{subfigure}
  \caption{Qualitative comparison of (a) pointwise cost, (b) in-focus surface area, and (c) selected focus distances for different methods. In (a), the pointwise focus distance cost, as defined in \eqref{eqn:cost function}, is visualized in the cool-warm color. In (b), the points captured in-focus are visualized in green, otherwise in red. In (c), the outer rings of transparent spheres are camera poses, while their focus points are visualized as black spheres.}
  \label{fig:method evaluation}
\end{figure}

Since the experiment on the simulation data does not consider the errors of the estimated camera poses and the estimated 3D shape, we further evaluate the methods on a TBP scan with a mannequin. As shown in Table \ref{t:evaluation}, our method outperforms the "Closest Focus" and the "Average Focus" by 17\% and 7\% in the total cost, and by 56.7\% and 15.1\% in the in-focus surface area respectively. Fig.~\ref{fig:method_evaluation_1} shows the qualitative comparison of the pointwise cost over the surface for the three methods.  Fig.~\ref{fig:method_evaluation_2} demonstrates the in-focus surface area. Fig.~\ref{fig:method_evaluation_3} illustrates the focus distances (represented as black spheres) determined by each method. As expected, "Closest Focus" effectively captures only body parts close to the cameras. Noting that the depth of field at each camera pose is particularly shallow when using the closest depth, due to the positive correlation between depth of field and focus distance. The "Average Focus" method, conversely, fails to appropriately capture body parts near the cameras. In contrast, our method adaptively selects focus distances based on the target's shape, resulting in improved quality. 
The visibility percentage of the surface is 98.9\%. The average system resolution across the surface is 0.0566 mm/pixel ($\sigma = $  0.0274 mm/pixel). 
\subsubsection{Qualititative evaluation of resulting image}
\begin{figure*}[thpb]
    \centering
  \includegraphics[width=0.95\textwidth]{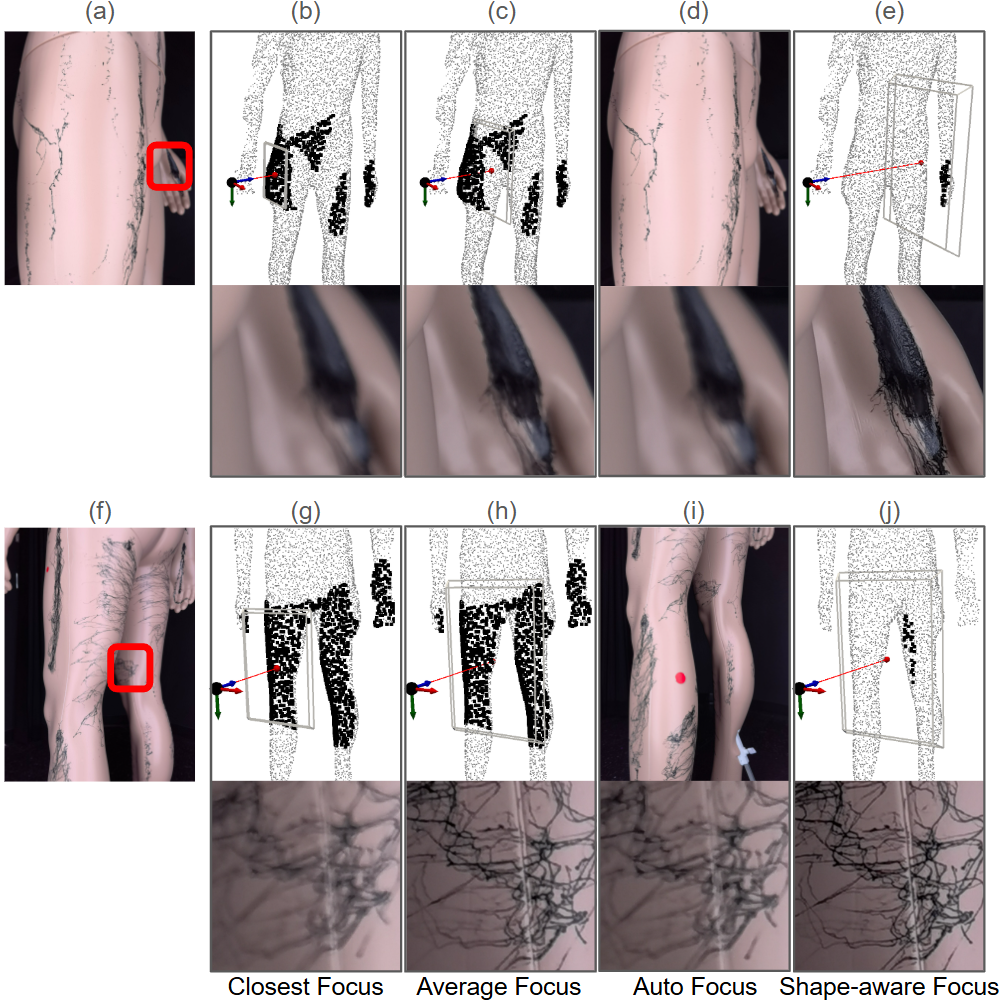}
  \caption{Qualitative comparison for body parts that are difficult to capture in focus. Two anatomical sites are evaluated: the palmar aspect of the left hand (a-e) and the medial aspect of the right knee (f-j). (a) and (f) display the camera perspectives, with red rectangles delineating the regions of interest. For the "Closest Focus" (b,g), "Average Focus" (c,h), and "Shape-aware Focus" (e,j) methods, the upper panels illustrate the camera pose (black sphere for camera position and colored axes for camera orientation), focus point (red sphere), and the view frustum bounded by the near and far depth of field limits. The lower panels show the resulting image crop for the regions of interest. In the upper panels of (b), (c), (g), and (h), black points represent surface areas visible to the camera, and thus are the set of points where the closest depth value and the average depth value are calculated for the camera. In the upper panels of (e) and (j), black points represent surface areas ``assigned'' to the camera by the shape-aware focus algorithm. For "Auto Focus" (d,i), the precise focus point is unknown. The upper panels of (i) and (j) show the view from the "Auto Focus" protocol that yields the highest image acuity for the region of interest. The black dots/lines in (a) and (f) are painted texture on the mannequin to compare the sharpness of the resulting images using different focus methods. The dots and lines do not affect the system’s performance because the proposed method does not rely on texture information. On the other hand, the dots and lines make it easier to focus successfully when using auto-focus. However, the proposed shape-aware focus still outperforms the auto-focus.
  \label{fig:qualitative 1}
  }
\end{figure*}

Since the goal of the proposed method is to improve the 2D image quality for TBP, we further evaluate our approach using the resulting images. We visualize the images captured by different methods for two body parts in Fig.~\ref{fig:qualitative 1}. Our method outperforms the baseline methods when capturing images of regions with a large variation in depth (within a single camera's field of view) such as the inner side of the arms, the hands, the knees, and the thighs. These regions are particularly difficult to capture in focus without using the proposed method.

\subsection{Robustness of Focus Distance Optimization}
As the optimized focus distances depend on the camera poses and 3D shape, we would like to assess their robustness with respect to noise in the camera calibration, noise in the surface estimation, as well as a slight change in body pose during a scan.
\subsubsection{Effect of camera calibration}
\label{ss:effect system calibration}
To evaluate robustness to camera calibration, we add normally distributed noise to the ground-truth camera parameters ($\mu\!=\!2.5\hbox{mm}\,/\,\sigma\!=\!1\hbox{mm}$ for translation and $\mu\!=\!0.3^\circ\,/\,\sigma\!=\!0.1^\circ$ for rotation, matching the estimated noise in \S\ref{ss:system precision}), optimize the focus distances using the noisy camera parameters, and measure the in-focus coverage when the focus distances computed using the noisy camera parameters are applied to the true camera parameters. 

Over 30 trials, an average of 0.06\% of the surface area was affected ($\sigma=0.16$\%), indicating that our system precision is sufficient for a robust estimation.

\subsubsection{Effect of 3D shape estimation}
\label{ss:effect 3D shape estimation}
To evaluate robustness to shape estimation, we use estimated 3D shapes as input for the shape-aware focus, derive focus distances for each camera, and evaluate the total cost on the ground-truth mesh using the computed focus distances. The resulting total costs are 2165.94 (Ground-truth mesh), 2255.68 (SMPL-NICP), and 2261.19 (TSDF). As expected, using estimated mesh results in a slightly increased cost compared to using the ground-truth mesh. However, the performance of both SMPL-NICP and TSDF methods is acceptable for practical applications, with only a 1\% decrease in surface coverage. 

\subsubsection{Effect of pose change}
\label{ss:effect pose change}
During the image capture, the patient needs to hold the same pose and avoid abrupt movements. However, a slight change in body pose is unavoidable. Among the small changes, the most significant one comes from the postural sway: small and continuous movements the body makes to maintain balance in quiet standing. The postural sway is reported to be within the range of 20 mm and 10 mm along the anterior-posterior and the medial-lateral direction for the center of mass of the body, respectively \cite{gatev1999feedforward}. To evaluate robustness to change in body pose, we fit an SMPL model to the original mesh (with \cite{zuo2021selfsupervised}), use the 3D shapes of the fitted mesh for the shape-aware focus, derive focus distances for each camera, and evaluate the total cost on a deformed mesh with changed pose. 

\begin{figure}
  \centering
  \begin{subfigure}{0.45\textwidth}
    \includegraphics[width=\textwidth]{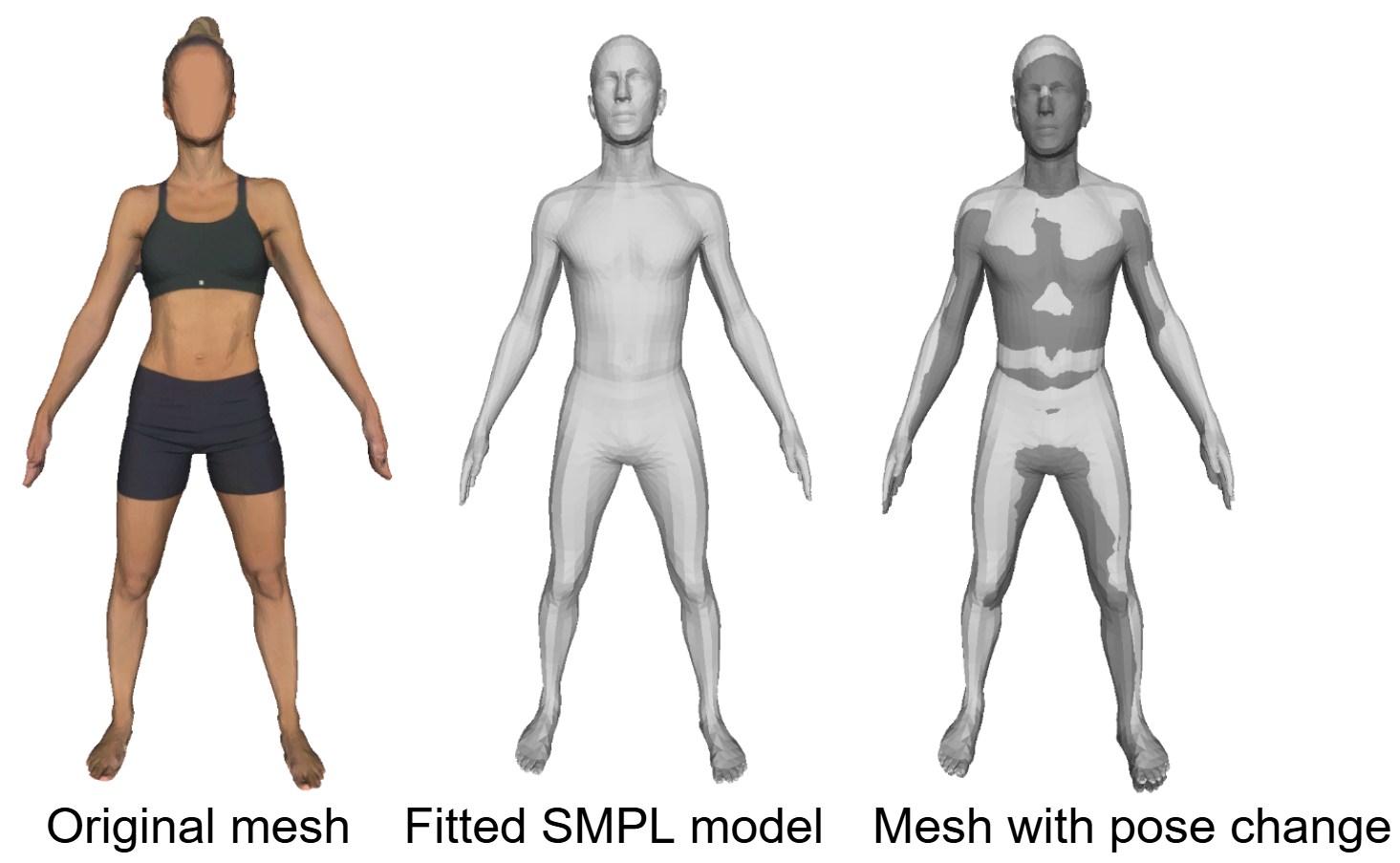}
    \caption{Visualization of simulation with pose change}
    \label{fig:pose data}
  \end{subfigure}
  \hfill
  \begin{subfigure}{0.45\textwidth}
    \includegraphics[width=\textwidth]{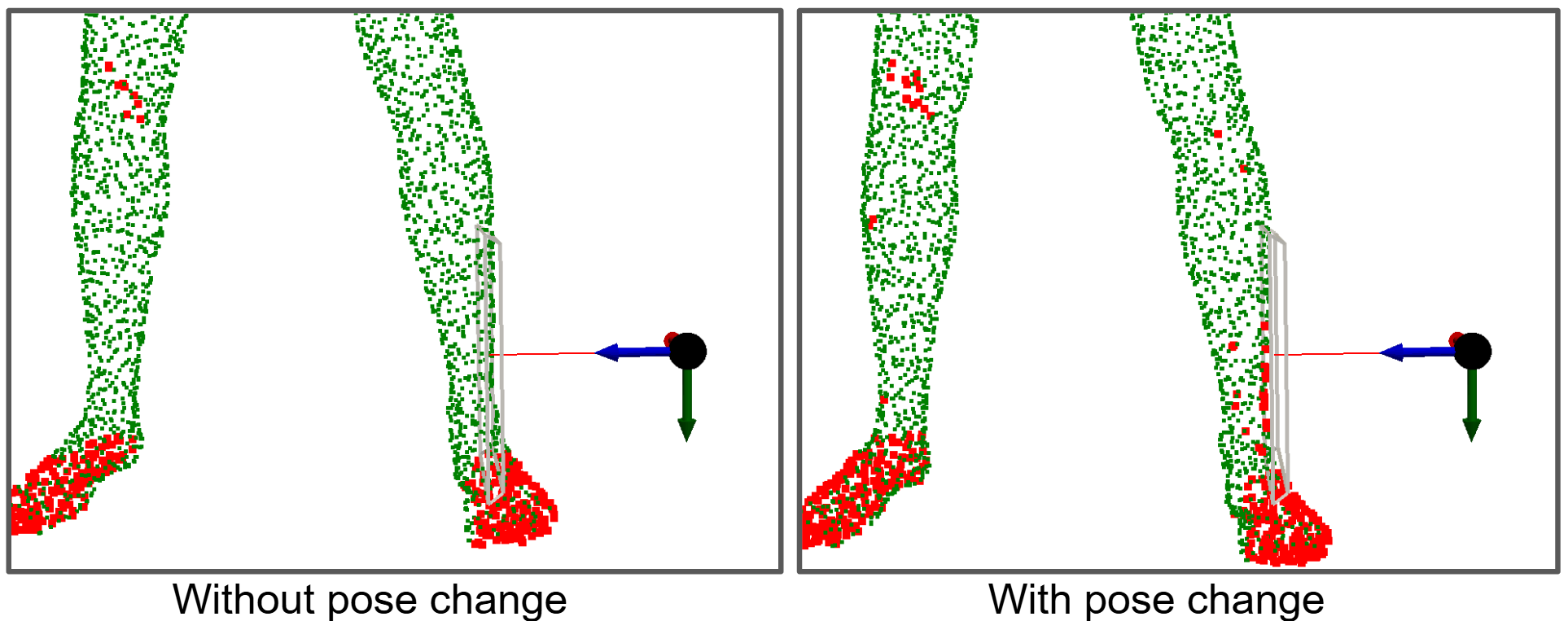}
    \caption{Visualization of the negative effect of pose change}
    \label{fig:pose effect}
  \end{subfigure}
  \caption{Visualization of (a) the simulation data with pose change and (b) the negative effect of pose change.
  (a) shows an original mesh from 3DBoyTex, the fitted mesh using a SMPL-based model, and the generated mesh of pose change to simulate body motions due to postural sway. In (b), the points captured in-focus are visualized in green, otherwise in red. A camera frame and its view frustum bounded by the near and far depth of field limits are plotted in the two sub-figures in (b). For the right sub-figure in (b), the red points next to the view frustum are the surface area that becomes out of focus due to the pose change.}
  \label{f:exp pose}
\end{figure}

We model the change in a patient's pose with the multi-joint coordination model \cite{krishnamoorthy2005joint,scholz2007motor}. We note that during the stabilization of body position, the oscillations of individual body segments typically do not exceed 1-2 degrees of joint movements \cite{ivanenko2018human}. Therefore, to generate a deformed mesh with changed pose, we apply a rotation of 2 degrees to six major body joints (ankle, knee, hip, lumbo-sacral junction, cervical spine, and atlanto-occipital) as used in \cite{hsu2007control}. We follow the axis-angle convention in SMPL to add a rotation to each of the six major body joints along a random axis by 2 degrees. The added rotations simulate large deviations the body pose could have during the scan. Fig.~\ref{fig:pose data} shows both the fitted SMPL model and the generated mesh with a slight change in pose for an example subject in 3DBdoyTex.

In our analysis, we exclude the meshes whose body segments are outside the volume of the scan and the meshes with self-intersection. Both cases are undesirable since they cause some body regions to be invisible to the system. In addition, both cases should be reasonably avoidable in total body photography. We evaluate the effect of pose change for 141 subjects in 3DBodyTex. For each subject, we create 10 poses of postural sway. The total costs are 2836 ($\sigma=246$\%) without pose change and 2939 ($\sigma=257$\%) with pose change, with a decrease of in-focus surface area by less than 2\%. Therefore, the proposed method is tolerant to slight changes in pose. Fig.~\ref{fig:pose effect} demonstrates the negative effect caused by a slight change in pose during a scan. We visualize the points that become out of focus due to the pose change. We observe that this could happen when a camera and its associated points are very close to each other because the depth of field is particularly shallow in this case.

\subsection{Operation Time}
The entire scanning process takes less than 100 seconds, including the following individual components: 1) Data collection for multi-frames of depth images (8 seconds). 2) TSDF-based 3D shape estimation (5 seconds). 3) Shape-aware focus computation (5 seconds). 4) Image capture (80 seconds). We note that the system calibration takes less than 10 minutes for both data collection and computation. However, the system calibration is performed only once after the camera configuration is determined and thus is not considered in the operation time. There is no significant difference in the in-focus surface area between SMPL-NICP and the TSDF method in \S\ref{ss:effect 3D shape estimation}. However, considering the processing time (120 seconds for SMPL-NICP and 5 seconds for TSDF), we adopt the TSDF method in the proposed pipeline.
Since the proposed method is tolerant to postural sway (\S\ref{ss:effect pose change}), the operation time is sufficiently short, assuming the patient can maintain the balance and remain relatively stationary in a natural pose (e.g. anatomical position).
As a comparison, the total scanning time is reported to be a few seconds (from \cite{ahmedt2023monitoring}), 3 minutes (from \cite{dengel2015total}), and 6 minutes (from \cite{korotkov2014new}) in the literature. While the scanning time of the system in \cite{ahmedt2023monitoring} is significantly shorter than ours, their system uses sixty cameras (as compared to seven cameras in our system) to allow for an instantaneous scan. We also note that a non-automated complete skin examination, with or without dermoscopy, takes less than 3 minutes on average \cite{zalaudek2008time}. Overall, we believe a scan within 100 seconds is reasonable for total body photography.

\section{Discussion}

\subsection{Cost, Operation Time, and Quality Trade-off in TBP System}
In Total Body Photography (TBP) system design, there is a trade-off among system costs, operational efficiency, and imaging data quality. The key determinants of system performance include, but are not limited to: the quantity and quality of cameras and lenses, system structure (robotic or static), scanning methodology (instantaneous capture or continuous scanning), system resolution, and image sharpness. Various focus protocols employed in different system designs may yield comparable imaging quality outcomes. For example, a static TBP system utilizing hundreds of cameras might effectively employ autofocus, assuming each camera is responsible for imaging a small, approximately planar portion of the body surface. However, such a system would likely incur significant costs.
While the optimal design of a TBP system remains a subject of ongoing research, we propose a cost-effective alternative that compensates for a small number of cameras by using a 360-degree rotational mechanism. Our continuous scanning approach prioritizes imaging quality over operation time. The proposed system establishes a benchmark for the cost-time-quality trade-off in TBP systems, facilitating quantitative comparisons and informing future system optimizations.

\subsection{Motion Blur}
Motion blur is the result of the relative motion between the camera and object during image capturing \cite{nayar2004motion}. There are two sources of motion blur in our setup. The first source comes from the camera movement during the image capture as the rotary beam is rotating around the patient. The second source comes from the postural sway as described in \S\ref{ss:effect pose change}. Therefore, we analyze the motion blur from the two sources separately.

To calculate the motion blur from camera movement, we estimate the number of pixels a point traverses during the exposure time when the camera is moving along a circular path around the patient. We first calculate the traversal in the number of degrees and then convert the traversed degrees to pixels based on the field of view (FOV) of the camera. Concretely, we have:
\begin{equation}
    \text{Motion Blur} = \frac{H}{\text{FOV}} \times \omega \times t,
\end{equation}
where
\begin{equation}
  \text{FOV} = 2\times\tan^{-1}\Bigl(\frac{\text{Sensor Dimension}}{2\times \text{Focal Length}}\Bigr).  
\end{equation}
% The camera’s shutter speed is set at 1/500 of a second. The sensor size is 1/1.55 inches. The focal length is 6 mm. The cameras are viewing toward the center axis of rotation. The radius of the cylindrical camera network is approximately 450 mm. The image resolution is 8192 x 6144.
Recall our camera settings in \S\ref{s:implementation}, during an exposure of 1/500 second, the angle traversed is 0.018$^\circ$ at 9$^\circ$/sec (the fastest speed in the setup). Since the rotation causes horizontal movement in the image, with a horizontal FOV of $47.66^\circ$, the resulting motion blur is 2.32 pixels. From our experiment with a mannequin, we do not observe the effect of motion blur due to camera movement in the resulting images.

To calculate the motion blur from the postural sway, we use the mean sway velocity of 13.42 mm/s, the resulting velocity from a velocity of 12 mm/s in the anterior-posterior direction and a velocity of 6 mm/s in the medial-lateral direction \cite{kim2010sex}. Considering that the radius of the cylindrical camera network is 450 mm, the angle traversed during exposure is 0.0034 degrees. The resulting motion blur is 0.44 pixels. As a comparison, we calculate the theoretical motion blur for the system proposed by Korotov et al. \cite{korotkov2014new}. The shutter speed is set at 1/30 second. The pixel size is 1.9 $\mu$m \cite{Korotkov2014}. The camera-to-subject distance is 400–450 mm. The image resolution is 4000 $\times$ 3000. The resulting motion blur from the postural sway in their system is theoretically 3.66 pixels. However, they concluded an acceptable tolerance to patient’s motion. In our system, the total motion blur of the proposed system is less than 3 pixels at the worst case when the camera is moving in the opposite direction of the postural sway. Therefore, we conclude that the motion blur has a negligible effect on the proposed system. In addition, the motion blur can be reduced by using a shorter exposure time with stronger illumination from the LED light panel.

\subsection{Image Navigation}
One of the byproducts of our formulation is that we can locate the best image a viewer software can display for any part of the body with minimal computation. We
compute the assignment function $\phi:\Mesh\rightarrow\Camera$ mapping each point on the surface to the camera that ``best'' images it. We take advantage of this by using the association to facilitate navigation of the acquired images -- allowing a clinician to select a point $p\in\Mesh$ on the surface and displaying the image from camera $\phi(p)$. Though image navigation has been informally proposed before (~\cite{korotkov2014new,ahmedt2023monitoring}), our approach is more effective, leveraging focal information to select the single best image when the point of interest is in the view frustum of multiple cameras.

\subsection{Limitations}
The proposed system and imaging pipeline are limited in several aspects. First, we do not consider the issue of focus breathing: a geometric change in the camera's field of view during focusing. Therefore, the points captured at the image boundary in the real world may differ in our model. However, we do not observe a significant effect on the change of the FOV when changing the focus distance (lens’s focus point) using a constant focal length lens in our system, partly because of the specific optical properties of our lens design. Additionally, the designed cost accounting for the deviation from the optical-axis (in Eq.~\ref{eqn:cost function}) penalizes points far from the image center. Therefore, focus breathing imposes an insignificant effect on our TBP configuration in practice. 
Second, a subject may wobble largely or undergo an abrupt change in body pose during a scan, resulting in a significant misalignment between the estimated 3D shape and the underlying 3D shape of the subject during the scan. Being able to stand still should be included in the inclusion criteria for any clinical use of the system. Compensation for the initial estimated 3D shape using deformable registration in real-time or shortening the scanning time may mitigate the misalignment. Alternatively, designing a recumbent system or adding a stabilizer for a patient to hold can reduce the misalignment as well.
Third, the proposed system and method do not optimize for image contrast. Therefore, the system may not be suitable for patients with different skin tones. To enhance image contrast, a future system may include light from different spectral sources and perform post-processing on images.
Fourth, though the proposed system is designed to accommodate patients with various shapes and poses, a patient must stay within the designated volume for a proper scan. Therefore, the system is not suitable for arbitrary poses, such as some of the poses used in 3DBodyTex dataset. However, the unsuitable poses in the dataset are also uncommon for total body photography.
Finally, some surface area may be occluded or captured at a relatively low resolution. In the conclusion section, we offer some future remedies to address this issue.

\section{CONCLUSION}
We propose a shape-aware TBP system to automatically capture full-body images while optimizing the image sharpness over the entire body surface. The imaging system consists of depth and RGB cameras on a 360-degree rotary beam. We develop a customized system calibration using incremental Structure from Motion. We estimate the 3D shape of the subject using the TSDF, which we show works better than the template-based method. We systematically evaluate the individual components of the proposed imaging pipeline and validate that the proposed shape-aware focus captures images that are sharper than the existing focus protocols. 

%Future work
In the future, the system can be extended to optimize for camera poses based on the 3D shape of the subject to improve the surface coverage and image quality. Alternatively, we can acquire multiple scans for a subject in different poses so that some commonly occluded body parts become visible, such as the armpit, scalp, and bottom of the feet. The method can be extended to optimize in-focus surface area and equivalent system resolution over multiple scans after registration. We will also add polarizing filters to the system if the skin glare is significant. We believe the high-fidelity imaging enabled by our system will improve automated skin lesion analysis for skin cancer screening.

% \bibliography{bibliography/references} % bibliography data
% \bibliographystyle{bibliography/IEEEtran} 
\bibliography{references} % bibliography data
\bibliographystyle{IEEEtran}

\end{document}